\documentclass[12pt, final]{l4dc2024}

\title[Residual Learning and Context Encoding]{Residual Learning and Context Encoding for Adaptive Offline-to-Online Reinforcement Learning}
\usepackage{times}

\usepackage{tikz}
\usepackage{pgfplots}
\usetikzlibrary{patterns}
\usetikzlibrary{decorations,fit,backgrounds,arrows.meta,calc}
\usetikzlibrary{shapes,arrows,positioning}
\pgfplotsset{every axis/.append style={
		legend style={inner xsep=1pt, inner ysep=0.5pt, nodes={inner sep=1pt, text depth=0.1em},draw=none,fill=none}
}}
\usepgfplotslibrary{groupplots}

\newlength{\figurewidth}
\newlength{\figureheight}
\newlength{\tblw}

\usepackage{xspace}
\newcommand{\eg}{\textit{e.g.\@}\xspace}
\newcommand{\ie}{\textit{i.e.\@}\xspace}

\usepackage{algorithm}
\usepackage{algpseudocode}

\coltauthor{\Name{Mohammadreza Nakhaei\textsuperscript{1}} \Email{mohammadreza.nakhaei@aalto.fi} \\
 \Name{Aidan Scannell\textsuperscript{1,2}} \Email{aidan.scannell@aalto.fi}\\
 \Name{Joni Pajarinen\textsuperscript{1}} \Email{joni.pajarinen@aalto.fi}\\
 \addr \textsuperscript{1}Department of Electrical Engineering and Automation, Aalto University, Espoo, Finland  \\
 \addr \textsuperscript{2}Finnish Center for Artificial Intelligence, Finland
 }

\usepackage{xcolor}         

\newcommand{\our}{\textsc{ReLCE}\xspace}

\usepackage[capitalise,nameinlink]{cleveref}
\crefname{section}{Sec.}{Secs.}
\crefname{algorithm}{Alg.}{Algs.}
\crefname{appendix}{App.}{Apps.}
\crefname{definition}{Def.}{Defs.}
\crefname{table}{Table}{Tables}
\begin{document}

\maketitle

\begin{abstract}%
 Offline reinforcement learning (RL) allows learning sequential behavior from fixed datasets. Since offline datasets do not cover all possible situations, many methods collect additional data during online fine-tuning to improve performance. In general, these methods assume that the transition dynamics remain the same during both the offline and online phases of training. However, in many real-world applications, such as outdoor construction and navigation over rough terrain, it is common for the transition dynamics to vary between the offline and online phases. Moreover, the dynamics may vary during the online fine-tuning. To address this problem of changing dynamics from offline to online RL we propose a residual learning approach that infers dynamics changes to correct the outputs of the offline solution. At the online fine-tuning phase, we train a context encoder to learn a representation that is consistent inside the current online learning environment while being able to predict dynamic transitions. Experiments in D4RL MuJoCo environments, modified to support dynamics' changes upon environment resets, show that our approach can adapt to these dynamic changes and generalize to unseen perturbations in a sample-efficient way, whilst comparison methods cannot\footnote{Code available at: \href{https://github.com/MohammadrezaNakhaei/ReLCE}{https://github.com/MohammadrezaNakhaei/ReLCE}}.

\end{abstract}

\section{Introduction}




Offline reinforcement learning (RL) \citep{levine_tutorial, survey} has the potential to learn policies to accomplish complicated tasks from offline data without interacting with the environment. 
However, the environment in which these policies are deployed can in practice differ from the environment where the data was collected. Therefore, a fine-tuning phase that makes the policies adaptive to different modifications to the environment is necessary for real-world applications. An example is hydraulic systems~\citep{egli2022general} where temperature influences the properties of the system and adaptation can be crucial.

Residual learning enables learning a residual agent that corrects the actions of a base policy. Prior research has used residual learning to combine conventional feedback controllers with RL agents \citep{residual_levine,residual_navigation, residual_race} for manipulation and navigation tasks, leading to improved sample efficiency. 
In this work, we use residual learning to adapt offline policies to environments with differing dynamics.

To enable the residual agent to adapt to the changes in the environment, we use a context encoder where a short history of previous transitions is used to infer the changes in the environment. We use the context encoder to learn a latent variable indicating the changes in the dynamics. Our representation learning relies upon minimizing the error over multi-step predictions as well as accurately predicting other transitions.

In this paper, the experimental focus is on MuJoCo \citep{mujoco} locomotion tasks from the D4RL \citep{d4rl} benchmark. 
We extend the environments such that during online fine-tuning we can resample transition dynamics’ parameters upon the environment reset at the start of each episode.
The assumption is that at the beginning of each episode, a new set of dynamics parameters is sampled from an unknown distribution. Once the dynamics parameters are sampled, the dynamics remain constant for the duration of an episode.
 
We show that our approach can adapt to different changes in dynamics parameters whilst considering the base offline policy and a short history of transitions. Further to this, we show that our approach generalizes to unseen dynamics parameters. That is, it generalizes to out-of-distribution changes in the environment that were not present during training. 

\section{Problem Statement}

In this paper, we consider offline-to-online RL. However, in contrast to previous approaches, we do not assume that the transition dynamics remain the same during the offline and online training phases. More specifically, we assume that during each episode of the online training phase, the transition dynamics are governed by one of $N_{M}$ sets of transition dynamics parameters.
The goal is to train an agent that can adapt to these dynamics changes using only a limited number of interactions.   


We assume that the offline dataset is collected from a \textit{Markov Decision Process} (MDP) \newline${M_{\text{offline}} = \langle S, A, R, P_{\text{offline}}, \gamma, \rho_0 \rangle}$ consisting of state space $S$, action space $A$, a scalar reward function $R$, transition dynamics $P_{\text{offline}}(s_{t+1}|s_t, a_t)$ which represent the distribution of possible next states conditioned on the current state and action, a discount factor $\gamma \in [0,1]$ and an initial state distribution $\rho_0(s_0)$. A single trajectory (episode) consists of the list of states, actions, and rewards ${\tau = [(s_1, a_1, r_1), (s_2, a_2, r_2), ..., (s_{T-1}, a_{T-1}, r_{T-1}), (s_T)]}$ where $s_T$ is the termination state. In online interactions, similar to previous meta-RL formulations, we assume a distribution of MDPs $p(M)$ with shared state space, action space, and reward function while the transition dynamics $P_i(s_{t+1} \mid s_t, a_t)$ vary between different MDPs. 
Each MDP is given by ${M_i = \langle S, A, R, P_i, \gamma, \rho_0 \rangle}$.
At the beginning of each trajectory in the online fine-tuning stage, an MDP $M_i$ is sampled from the distribution $p(M)$ and is consistent until the next trajectory. The objective is to find the policy $\pi$ that maximizes the expected cumulative reward
\begin{equation}
  J(\pi) = \mathbb{E}_{M_i \sim p(M), s_0 \sim \rho_0(s_0), s_{t+1} \sim P_i( s_t, a_t), a_t \sim \pi}  \left [\sum_{t=0}^T \gamma^t r(s_t,a_t) \right] \;.
\end{equation}


\section{Related Work}
In this paper, we propose a novel challenge at the intersection of offline-to-online RL and adaptive RL. In this section, we first present an overview of offline RL methods since we use an offline policy as the base policy in the residual learning framework. Then we discuss different approaches for offline-to-online RL and compare them to our setting. Finally, we describe previous research on adaptive RL and illustrate how our approach is different and unique.

\subsection{Offline Reinforcement Learning} 
Offline RL algorithms try to learn a policy that maximizes expected cumulative reward while only using static datasets without further interaction with the environment. The challenge in offline RL is distribution shift where the learned policy deviates from the behavior policy (the policy used to collect data) and selects out-of-distribution (OOD) actions for bootstrapping. To mitigate this, several methods constrain the policy to stay close to the behavior 
policy \citep{bcq,bear,brac,abm_mpo,td3bc}. Another solution is to train pessimistic value functions \citep{cql,mopo,morel,combo,edac,cbop,mcq,csve} where regularization is applied to penalize the action value functions for OOD actions. Another class of methods uses in-sample learning \citep{awr,awac,iql,idql,sql} where only the actions in the datasets are considered for training value functions/policy and perform weighted imitation learning on the behavior policy. Other methods use conditional generative modelling \citep{dt,tt,odt,diffuser,dd} to learn policies from offline datasets, sidestepping the need for bootstrapping and learning value functions. 

\subsection{Offline-to-Online Reinforcement Learning}
Pre-training on large datasets followed by fine-tuning on down-stream tasks has been investigated in modern machine learning, \eg, computer vision  \citep{pre_train_cv1,pre_train_cv2,pre_train_cv3} and natural language processing (NLP) \citep{pre_train_nlp1,pre_train_nlp2,pre_train_nlp3}.  To improve the performance of well-trained offline policies, \citet{awac} imitated actions with high advantage estimates, \citet{balanced_replay} modified the sampling method to incorporate near on-policy offline data, \citet{adaptive_bc} carefully adjusted the behavior cloning regularization weight, and \citet{q_ensemble_adaptation,robust_ensemble_adaptation} used an ensemble of value functions whilst considering uncertainty and smoothness. Recently \citet{pex} proposed using policy expansion sets where an offline policy is fixed and new policies are trained; adaptive composition of policies is used to interact with the environment. This work is close to our method since the offline policy is fixed during online fine-tuning. Still, all these methods assume that the environment during online adaptation is fixed and consistent with the offline dataset. In contrast, our method learns an adaptive policy during online fine-tuning. Hybrid RL \citep{hybrid_rl} considers imperfect simulators with offline data from the real environment and uses both to learn a policy, but the learned policy is not adaptive. 

\subsection{Adaptive Reinforcement Learning}
Adaptive RL aims at improving the generalization of policies across dynamic changes and different tasks. \citet{osi} propose to learn an online system identification module using supervised learning and then train a universal policy considering the predicted system parameters. \citet{rma,rma_bipedal} instead, use a simulator while varying different parameters and learn an adaptation module by predicting the latent space of system parameters from a history of transitions. \citet{mrac,l1} incorporate adaptive control to estimate and compensate for changes in the dynamics, these methods require a nominal dynamic model and are restricted to the environment with Lagrangian mechanics without contacts. Meta-RL methods based on \textit{Model Agnostic Meta Learning} (MAML) \citep{maml} learn a pre-trained model from a set of environments and adapt to a new environment within several updates \citep{meta_model_base}. Methods based on memory use past interactions to update the policy. In \textit{PEARL} \citep{pearl}, a context encoder is learned from previous transitions to facilitate learning the action value function. \citet{cadm} proposed \textit{Context-aware Dynamic Model} (CaDM) to learn a latent vector from previous transitions and use the latent space in the dynamic model for more accurate predictions. \citet{tmcl} incorporate multiple choice learning in learning context-aware dynamic model. In \citet{context_everything}, N random transitions ($s,a,s^{\prime}$) from a single environment (trajectory) are passed through the encoder to infer the context accordingly. These works are similar to our method in the regard that prediction error is used to train the context encoder, but our method considers multi-step prediction loss, and future/past prediction loss, and also uses similarity loss to encourage consistency of the latent space in the same environment. In addition, our framework considers an offline fixed policy as the base policy and trains the residual agent on top of that.




\section{Method}
\begin{figure}[ht!]
\centering
\includegraphics[width=1 \textwidth]{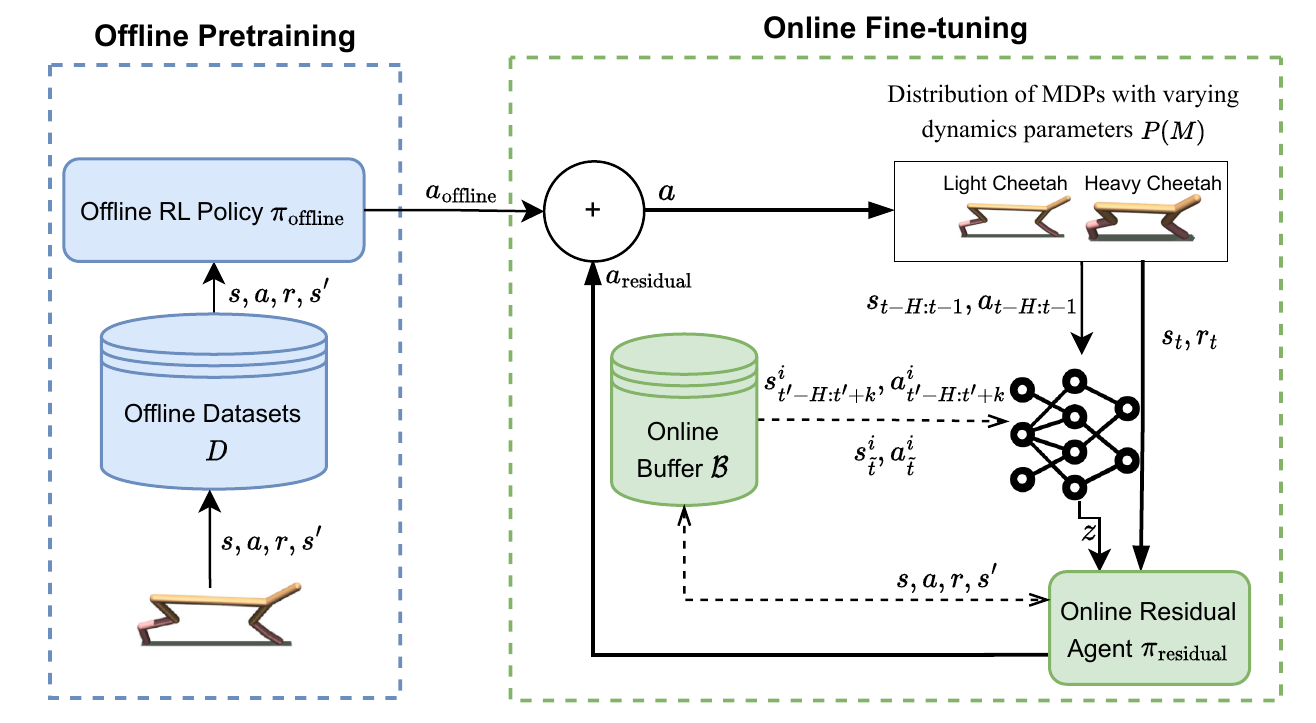}
\caption{\textbf{\our overview} - The offline policy $\pi_{\text{offline}}$ is used as base policy trained on existing datasets $\mathcal{D}$. The context encoder infers the changes in the environments, and the residual agent compensates for the modifications by considering the context and offline policy.} 
\label{fig:overview}
\end{figure}

Given an offline policy $\pi_{\text{offline}}$, we propose an adaptive policy that learns a residual policy $\pi_{\text{residual}}$ to account for errors in the offline policy. Our method consists of an offline agent, context encoder/decoder, and residual agent,
\begin{align} 
\label{eq:total_action} a_t &= \alpha \pi_{\text{offline}}(a^{\text{offline}}_t \mid s_t) + (1-\alpha)\pi_{\text{residual}}(a^{\text{residual}}_t \mid s_t, a^{\text{offline}}_t, z_t) \\
    \label{eq:encoder} z_t &= e_\theta (s_{t-H:t-1}, a_{t-H:t-1}) \\
    \label{eq:decoder} \hat{s}_{t+1} &= d_\theta (s_{t}, a_{t}, z_t), \;
\end{align}
where $z_t \in R^d$ is a context variable, \ie a latent variable indicating which MDP we believe we are in, $\alpha$ is a mixing coefficient, $e_\theta$ is the context encoder summarizing previous transitions, and $d_\theta$ is the decoder that predicts future states used for learning representations. In the remainder of this section, we discuss each aspect of our method in more detail. \cref{fig:overview} provides an overview of our method and \cref{alg:finetune} summarizes the online fine-tuning procedure.

\begin{algorithm}
\caption{Adaptive online fine-tuning with residual agent}\label{alg:finetune}
\textbf{Require}: trained offline agent $\pi_{\text{offline}}$, distribution of MDPs $p(M)$, context length $H$ \newline 
\textbf{Initialize} context encoder $e_\theta$, decoder $d_\theta$, residual agent $\pi_{\text{residual}}$ and replay buffer $\mathcal{B}$ \newline
\For{step in training steps}{
    Sample an MDP from the distribution $M_i \sim p(M)$ \newline
    Set $t \gets 0$ and observe the initial state $s_0$ \newline
    Initialize $S_{\text{episode}} = \{ s_0 \}, A_{\text{episode}} = \{ \}, R_{\text{episode}} = \{ \} $ \newline
    \While{not done}{
        \eIf{$t<H$}{$a_t \sim \pi_{\text{offline}}(a_t|s_t)$}{
             Compute context $z_t$ according to  \cref{eq:encoder} \newline
             Compute the state of the residual agent $s^{\text{residual}}_t$ according to \cref{eq:s_residual} \newline
             Get the total action $a_t$ according to \cref{eq:total_action}}
        Interact with the environment with action $a_t$, observe the new state $s_t^\prime$ and the reward $r_t$ \newline 
        Add new state $s_t^\prime$, action $a_t$, and reward $r_t$ to $S_{\text{episode}}$, $A_{\text{episode}}$, $R_{\text{episode}}$ and set $t \gets t+1$ 
        
        Sample training batch $\{ s_{t^\prime-H:t^\prime}^i, a_{t^\prime-H:t^\prime}^i, s_{\tilde{t}}^i, a_{\tilde{t}}^i, r_{\tilde{t}}^i \}$ from the buffer $\mathcal{B}$ \newline
        Train context encoder using objective in \cref{eq:context_encoder} and train the residual agent using \textit{SAC}
    }
    Add trajectory ($S_{\text{episode}}, A_{\text{episode}}, R_{\text{episode}}$) to the buffer $\mathcal{B}$
}
\end{algorithm}


\textbf{Offline Agent:}
We consider \textit{Conservative Offline Model-Based policy Optimization} (COMBO) \citep{combo} for training the offline policy $\pi_{\text{offline}}$. This algorithm extends \textit{Conservative Q-Learning} (CQL) \citep{cql} by using samples from the learned ensemble of a probabilistic dynamic model to train a less conservative Q function. We used the same hyper-parameters as the original paper and trained the agent for one million gradient steps.

\textbf{Context Encoder:}
We train the context encoder to infer the changes in the environment from a short history of previous transitions implicitly without knowing the dynamic parameters. To train the context encoder, we use the forward dynamic prediction error. The decoder takes the latent representation $z_t$, state $s_t$, and action $a_t$ and predicts the next state $s_{t+1}$. To encourage learning a representation for long horizon predictions, we use a $k$ step loss function, where we use the predicted state to make further predictions. We also consider predicting other transitions from the same trajectory. This enables the encoder to learn a representation that is useful for prediction along the same trajectory (same MDP). The transition is randomly selected from the same trajectory. Prediction objectives do not guarantee consistency of the latent space along a trajectory where the assumption is that during the online fine-tuning, dynamics changes occur along episodes. To address this issue, we add an objective to maximize the cosine similarity loss between latent vectors of the same environment. From the same trajectory in memory, we sample \textit{N} sequence of transitions and compute latent vectors for each of them. We combine these objectives to obtain our encoder's objective:
\begin{multline} \label{eq:context_encoder}
    \mathcal{L}(\{s_{t_i-H:t_i+k}^i, a_{t_i-H:t_i+k}^i, s_{t^\prime_i}^i, a_{t^\prime_i}^i, s_{t^\prime_i+1}^i \}_{i=0}^N;\theta) 
    = \underbrace{\sum_{i=1}^N \sum_{k=0}^{K-1} \gamma^k ||d_\theta(\hat{s}_{t_i+k}^i, a_{t_i+k}^i, z_{t_i}^i)-s_{t_i+k+1}^i||^2}_{k\text{-step predictions}} +
    \\ 
     \underbrace{\sum_{i=1}^N||d_\theta(s_{t^\prime_i}^i, a_{t^\prime_i}^i, z_t^i)_-s_{t^\prime_i+1}^i||^2}_{\text{future/past prediction}} +
    \underbrace{\frac{\omega}{N-1}\sum_{i,j=1, i\ne j}^N -\frac{{z_{t_i}^i}^T}{||{z_{t_i}^i}||_2} \frac{{z_{t_j}^j}}{||{z_{t_j}^j}||_2}}_{\text{consistency}}   ,
\end{multline}
where $d_\theta$ is the decoder which is trained along with the encoder, $\hat{s}$ represents the predicted states, $\omega$ is a hyper-parameter to balance consistency and prediction objectives. In the first step, we use the actual state to make the predictions \ie $\hat{s}_t^i = s_t^i$.



\textbf{Residual Agent:}
The residual agent aims to learn an adaptive policy that maximizes the expected cumulative reward across the distribution of MDPs by online interaction with the environments. At the beginning of each episode, we modify the dynamics by sampling from the distribution of MDPs. The transition dynamics then remain fixed until episode termination.
We use a context encoder to infer the environment from a short history of previous transitions according to \cref{eq:encoder}. The residual agent observes the state of the environment, context vector, and the action of the offline policy,
\begin{equation} \label{eq:s_residual}
    s^{\text{residual}}_t = [s_t, a^{\text{offline}}_t, z_t]^T \;.
\end{equation}
We train the residual agent to compensate for the offline policy and output corrective actions. In \cref{eq:total_action}  $\alpha$ is a hyper-parameter with default value $\alpha = 0.75$ that chooses the importance of the
offline policy vs.\ the residual policy.

We use the offline policy until the time step is more than the sequence length of the context encoder, \ie $t=H$. Then we compute the context vector according to the prior transitions and determine the state of the residual agent according to \cref{eq:s_residual}. For training the residual agent, we use the \textit{Soft Actor Critic} (SAC) algorithm~\citep{sac}. The context encoder is fixed during the optimization of the Q-functions and the policy.  

\section{Experiments}


We evaluate our approach on continuous control tasks with different datasets from the D4RL~\citep{d4rl} benchmark. At the beginning of each episode, the mass of each link and the damping ratio of each joint are scaled by random numbers sampled from $[0.75, 0.85, 1, 1.15, 1.25]$ uniformly. We aim to answer the following questions:
\begin{itemize}
    \item Can our context-aware residual agent learn to adapt to transition dynamics changes?
    \item Can our context encoder learn representations that enable the agent to predict future states across different dynamics parameters? 
    \item Can our approach generalize to transition dynamics not seen during training?
\end{itemize}

\subsection{Evaluation of the Adaptation Performance}
In this section, we try to answer the first question and compare our methods adaptation performance to the baselines. The aim is to learn an adaptive policy during online fine-tuning with a limited number of interactions. We consider the following baselines:
\begin{itemize}
    \item \textbf{Recurrent SAC} \citep{sac_recurrent} uses recurrent networks for policy and value functions. We consider two variations: in the first variation we train the agent from scratch only by interacting online. In the second variation, we use residual learning with an offline policy similar to our method. We consider this baseline since it directly uses a history of transition to learn the policy and value function. 
    In contrast, our method infers the dynamics' context $z$ and trains an adaptive policy conditioned on this context. 
    
    \item \textbf{Meta RL} algorithms learn an adaptive policy from a set of environments. \textbf{PEARL} \citep{pearl} is an off-policy meta RL algorithm that includes a probabilistic context encoder. The comparison to this baseline can demonstrate the effect of offline policy and decoupling training the context encoder and policy learning.

    \item To demonstrate the necessity of adaptive online fine-tuning, we compare to \textbf{offline-to-online} methods. We consider \textbf{PEX} \citep{pex} and \textbf{Adaptive BC} \citep{adaptive_bc} for comparison. 
\end{itemize}
For all the baselines, we use official implementations with our distribution of MDPs. We use one-dimensional \textit{Convolution Neural Network} (CNN) to capture the temporal correlation between samples with $[4,2,1]$ kernel size followed by \textit{ReLU} activation function. Convolution layers are followed by a linear layer that outputs the latent vector. For the decoder, we use \textit{Multi-Layer preceptron} (MLP) networks with $[256, 256]$ hidden layers and \textit{ReLU} activations. We also use Adam optimizer to train the encoder and decoder with a learning rate of $0.0001$ while normalizing the target predictions. We use a default sequence length of $H=10$, latent dimension of $8$, and $K=5$ step prediction loss for training the context encoder. We use $N=4$ trajectories from the same environment and set $\omega$ to 0.1 to balance consistency and prediction objectives. We use MLP networks with $[256, 256]$ hidden layers followed by \textit{ReLU} activations for the actors and critics networks of the residual agent. We use the Adam optimizer with learning rates of $0.0001$ and $0.0003$ to train actor and critic networks respectively. 

We summarize the results in \cref{tab:result}. Directly using recurrent networks in \textit{SAC} without considering the offline agent has the worst performance and is not sample-efficient. Residual learning with the offline policy as the base improves the performance and sample-efficiency of \textit{SAC} with recurrent networks. This suggests that a residual framework using the offline policy simplifies the problem and increases sample efficiency. \textit{Residual RNN SAC} has a competitive performance in the \textit{hopper} environment for different types of datasets, however, for other environments, it cannot learn the task within 250k time-steps. \textit{PEARL} is more sample-efficient than \textit{RNN SAC} and even has a better performance than \textit{Residual RNN SAC} in \textit{halfcheetah}, but within the sample budget, it cannot learn to adapt to different changes in the dynamics effectively. In the online fine-tuning phase, \textit{PEX} and \textit{Adaptive BC} improve performance when interacting with modified environments and learn more robust policies compared to the offline agent. Surprisingly, \textit{Adaptive BC} outperforms our method without considering any context or history in the \textit{hopper} environment. We speculate that with different dynamic perturbations (changes in the mass and damping ratio) in the environment, the agent learns to behave conservatively and applies more torque/force trying to jump and move forward for different masses. To evaluate our speculation, we consider \textit{invertedpendulum} environment and we scale the mass from $[0.25, 0.5, 0.75, 1, 1.5, 2, 3, 5]$ uniformly. We collect the dataset for the offline agent by training \textit{SAC} for 200k timesteps and we use the replay buffer for the dataset. This environment is sensitive to the mass and simply applying more force/torque in different masses is not a solution. Methods that consider history including ours outperform offline-to-online methods demonstrating the necessity for adaptation.

\begin{table}[]
\scriptsize
	\renewcommand{\arraystretch}{1.}
	\setlength{\tabcolsep}{4.pt}
	\setlength{\tblw}{0.06\textwidth}

	\newcommand{\val}[2]{%
		$#1$\textcolor{gray}{\tiny ${\pm}#2$}
	}
    \centering
    \begin{tabular}{l|cccccc}
    \hline
         {\sc task} &  {\sc rnn sac} & {\sc residual rnn sac} & {\sc pearl} & {\sc pex} & {\sc adaptive bc} & \our (Ours)\\ \hline
        { hopper-medium-v2} &  \val{35.64}{10.04} & \val{65.24}{12.03} & \val{43.37}{8.03} & \val{69.98}{28.83} & \val{102.55}{0.99} & \val{94.74}{0.88} \\
        { hopper-medium-replay-v2} & \val{35.64}{10.04} & \val{86.13}{9.72} & \val{43.37}{8.03} & \val{85.18}{21.86} & \val{107.8}{2.26} & \val{97.70}{0.70} \\
        { hopper-expert-v2} & \val{35.64}{10.04} & \val{97.46}{5.36} & \val{43.37}{8.03} & \val{89.39}{23.97} & \val{108.09}{1.80} & \val{97.90}{0.64} \\ \hline
        { halfcheetah-medium-v2} & \val{9.22}{0.69} & \val{40.58}{4.28} & \val{64.65}{8.03}  & \val{62.98}{3.08} & \val{83.40}{3.35} & \val{92.86}{2.50} \\
        { halfcheetah-medium-replay-v2} & \val{9.22}{0.69} & \val{43.14}{1.73} & \val{64.65}{8.03} & \val{53.67}{1.70} & \val{78.88}{3.19} & \val{93.93}{3.47} \\
        { halfcheetah-expert-v2} & \val{9.22}{0.69} & \val{59.15}{8.84} & \val{64.65}{8.03} & \val{91.6}{2.43} & \val{85.42}{3.41} & \val{95.87}{3.22}\\ \hline
        { walker2d-medium-v2} & \val{13.00}{2.64} & \val{32.83}{8.60} & \val{16.63}{7.11}  & \val{80.60}{16.53} & \val{72.96}{8.60} & \val{90.80}{3.66} \\
        { walker2d-medium-replay-v2} & \val{13.00}{2.64} & \val{51.76}{9.46} & \val{16.63}{7.11} & \val{87.50}{9.98} & \val{96.15}{4.08} & \val{92.82}{2.04} \\
        { walker2d-expert-v2} & \val{13.00}{2.64} & \val{47.97}{5.49} & \val{16.63}{7.11} & \val{96.15}{9.78} & \val{103.86}{3.26} & \val{104.37}{3.04} \\ \hline
        { invertedpendulum-replay} & \val{92.82}{7.11} & \val{98.17}{3.32} & \val{74.73}{16.85}& \val{76.40}{7.20} & \val{72.67}{9.57} & \val{100}{0} \\
    \end{tabular}
    \caption{Results for adaptive online fine-tuning after 250k time-steps averaged over 10 random seeds, scores are normalized according to D4RL.}
    \label{tab:result}
\end{table}

\begin{figure}[!htbp]
\centering
\pgfplotsset{axis on top,scale only axis,width=\figurewidth,height=\figureheight, ylabel near ticks,ylabel style={yshift=-2pt, font=\small},y tick label style={rotate=90},legend style={nodes={scale=1., transform shape}},tick label style={font=\small,scale=1}, xlabel style={font=\small}, title style={font=\small}}
  \pgfplotsset{xlabel={Time step}, axis line style={rounded corners=2pt}}
  \pgfplotsset{axis line style={rounded corners=2pt}}
  \setlength{\figurewidth}{.26\textwidth}
  \setlength{\figureheight}{.55\figurewidth}

    \input{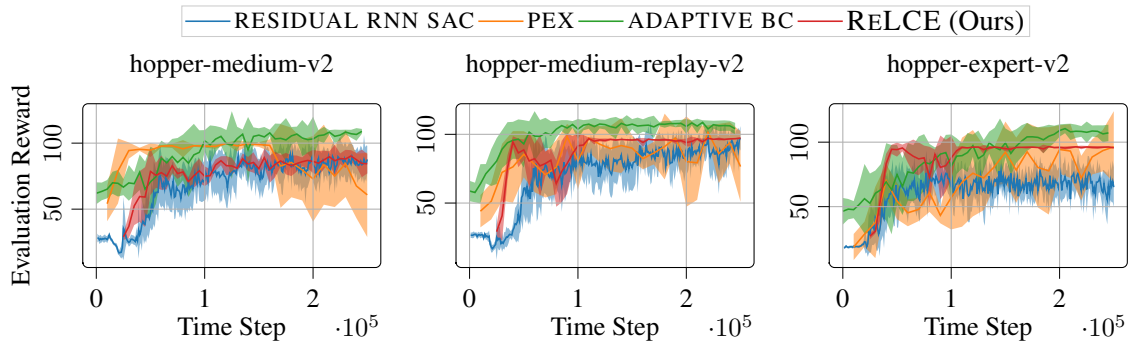}
    \caption{\textbf{Learning curves in hopper environment} Our method (red) is most sample efficient (\ie converges in few environment steps) when the offline policy is learned with the expert data set (right) but still performs well with the medium data set (left). The shaded regions represent the standard deviation over 10 random seeds.}
\label{fig:learning_curves}
\end{figure}

\cref{fig:learning_curves} shows the learning curves for different methods for the \textit{hopper} environment with different datasets. In the methods that use residual learning, there is a drop in performance at the initial stage of fine-tuning since the residual agent is initialized and still not trained. 

\subsection{Latent Space Evaluation}
\begin{figure}[!htbp]
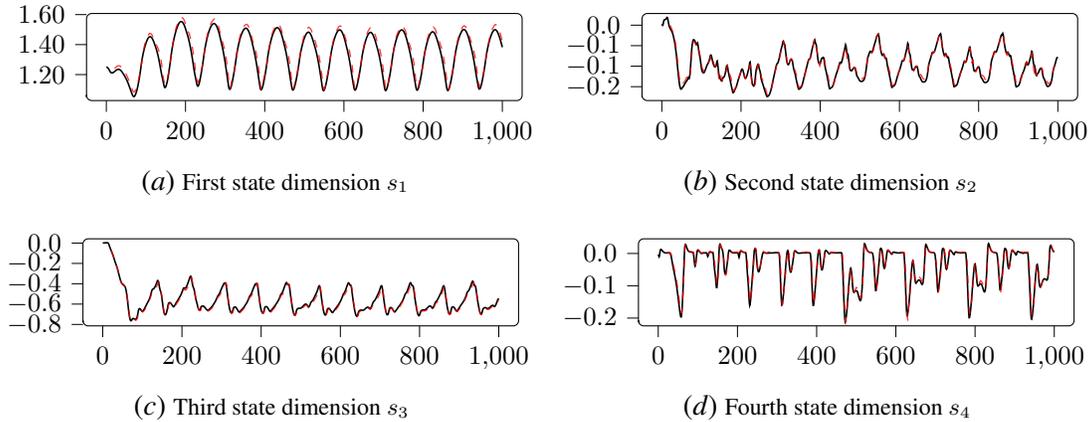

\centering
  \pgfplotsset{axis on top,scale only axis,width=\figurewidth,height=\figureheight, ylabel near ticks,ylabel style={yshift=-2pt, font=\small},y tick label style={font=\small},legend style={nodes={scale=1., transform shape}},tick label style={font=\small,scale=1}, xlabel style={font=\small}, title style={font=\small}}
  \pgfplotsset{axis line style={rounded corners=2pt}}
  \pgfplotsset{
      y tick label style={
        /pgf/number format/.cd,
        fixed,
        fixed zerofill,
        precision=1,
        /tikz/.cd
    }
  }
  \setlength{\figurewidth}{.38\textwidth}
  \setlength{\figureheight}{.2\figurewidth}
  
  \pgfplotsset{grid style={dotted},title={\empty}, scale only axis}

    \subfigure[\footnotesize First state dimension $s_1$]{  \pgfplotsset{
      y tick label style={
        /pgf/number format/.cd,
        fixed,
        fixed zerofill,
        precision=2,
        /tikz/.cd
    }
  }\input{imgs/tex_files/predict_10steps_0}}
    \subfigure[\footnotesize Second state  dimension $s_2$]{\input{imgs/tex_files/predict_10steps_1}}
    \subfigure[\footnotesize Third state dimension $s_3$]{\input{imgs/tex_files/predict_10steps_2}}
    \subfigure[\footnotesize Fourth state dimension $s_4$]{\input{imgs/tex_files/predict_10steps_3}}
  \caption{\textbf{Multi-step predictions} \our makes accurate state predictions (10 steps) using the context encoder (red dashed), when compared to the ground truth (black).}
  \label{fig:predictions}
\end{figure}

To evaluate the context encoder, we consider 10-step state predictions according to the latent space and the decoder. \cref{fig:predictions} shows predictions along a single trajectory for a randomly modified \textit{hopper} environment. The 10-step future predictions for different states of the environment are close to the observed state even though we use 5 future steps for training. This indicates that the learned representation by the encoder can infer the changes in the environment.

\begin{figure}[!b]
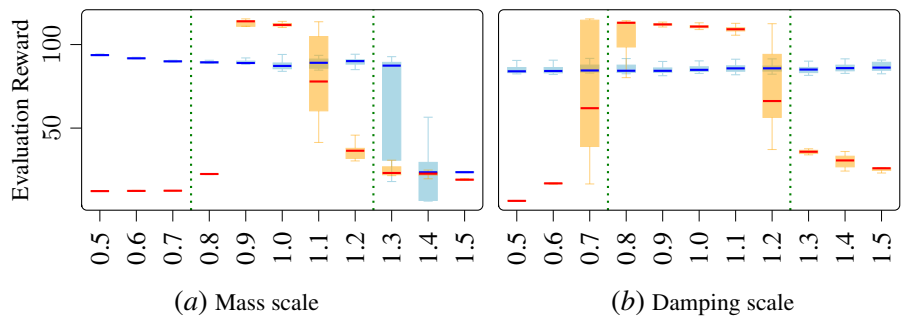

\centering
  \pgfplotsset{axis on top,scale only axis,width=\figurewidth,height=\figureheight, ylabel near ticks,ylabel style={yshift=-2pt, font=\footnotesize},y tick label style={rotate=90},  x tick label style={rotate=90}, legend style={nodes={scale=1., transform shape}},tick label style={font=\small,scale=1}, xlabel style={font=\small}, title style={font=\small}}
  \pgfplotsset{axis line style={rounded corners=2pt}}
  \setlength{\figurewidth}{.35\textwidth}
  \setlength{\figureheight}{.5\figurewidth}
  
  \pgfplotsset{grid style={dotted},title={\empty}, scale only axis}

    \subfigure[\footnotesize Mass scale]{\pgfplotsset{xlabel={}, xlabel style={yshift=-5pt},  xtick={1,2,3,4,5,6,7,8,9,10,11}, xticklabels={0.5, 0.6, 0.7, 0.8, 0.9, 1.0, 1.1, 1.2, 1.3, 1.4, 1.5}, ylabel={Evaluation Reward}}\input{imgs/tex_files/OOD_mass}}
    \subfigure[\footnotesize Damping scale]{\pgfplotsset{xlabel={}, xlabel style={yshift=-5pt},  xtick={1,2,3,4,5,6,7,8,9,10,11}, xticklabels={0.5, 0.6, 0.7, 0.8, 0.9, 1.0, 1.1, 1.2, 1.3, 1.4, 1.5}, ymajorticks=false}\input{imgs/tex_files/OOD_damp}}

    \caption{\textbf{Performance in the presence of dynamics changes} - Our method (blue) maintains good performance over a wider range of dynamics parameters than \textit{Adaptive BC} (red) on the hopper-medium-replay task. Green lines separate in-distribution and out-of-distribution changes. Boxes represent the interquartile range (IQR) with median.}
  \label{fig:indist_boxplots}
\end{figure}

\subsection{Generalization to Unseen Dynamics}

In this section, we investigate if our approach can adapt to changes in the environment that were not included during online fine-tuning. First, we investigate the same type of changes but with different magnitudes. \cref{fig:indist_boxplots} shows how different mass and damping ratios affect our method's performance, in terms of evaluation reward. We evaluate our method and \textit{Adaptive BC} 100 times for each change in the environment. While \textit{Adaptive BC} has a better performance for changes included in the training, especially small changes, our method can generalize to out-of-distribution changes. We speculate that for higher values of mass, the limitation in the actions (torques) makes it impossible for the agent to perform the task. 

\begin{figure}[!t]
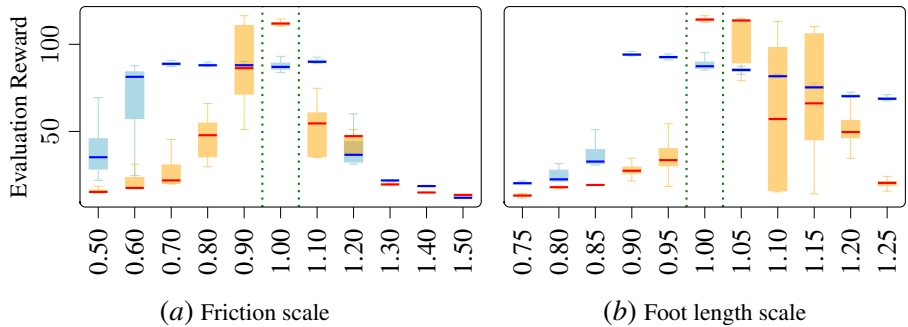

\centering
  \pgfplotsset{axis on top,scale only axis,width=\figurewidth,height=\figureheight, ylabel near ticks,ylabel style={yshift=-2pt, font=\footnotesize},y tick label style={rotate=90},legend style={nodes={scale=1., transform shape}},tick label style={font=\small,scale=1}, xlabel style={font=\small}, title style={font=\small}}
  \pgfplotsset{axis line style={rounded corners=2pt}}
  \setlength{\figurewidth}{.35\textwidth}
  \setlength{\figureheight}{.5\figurewidth}
  
  \pgfplotsset{grid style={dotted},title={\empty}, scale only axis}

    \subfigure[\footnotesize Friction scale]{\pgfplotsset{xlabel={}, xlabel style={yshift=-10pt}, xtick={1,2,3,4,5,6,7,8,9,10,11}, xticklabels={0.50, 0.60, 0.70, 0.80, 0.90, 1.00, 1.10, 1.20, 1.30, 1.40, 1.50}, x tick label style={rotate=90}, 
    ylabel={Evaluation Reward}}\input{imgs/tex_files/OOD_friction}}
    \subfigure[\footnotesize Foot length scale]{\pgfplotsset{xlabel={}, xlabel style={yshift=-10pt}, xtick={1,2,3,4,5,6,7,8,9,10,11}, xticklabels={0.75, 0.80, 0.85, 0.90, 0.95, 1.00, 1.05, 1.10, 1.15, 1.20, 1.25}, x tick label style={rotate=90}, ymajorticks=false} \input{imgs/tex_files/OOD_foot}}

    \caption{\textbf{Generalization to out-of-distribution dynamics parameters} - Our method (blue) shows some generalization to dynamics parameters outside of the distribution used during the online training on hopper-medium-replay. In contrast, \textit{Adaptive BC} (red) struggles to generalize outside the training distribution. Green lines separate in-distribution and out-of-distribution changes. Boxes represent the interquartile range (IQR) with median.}
  \label{fig:outdist_boxplots}
\end{figure}

Next, we consider different dynamics changes that were not used in training. To this end, we change the friction coefficients for the joints and the foot length. \cref{fig:outdist_boxplots} represent the results for our method and \textit{Adaptive BC}. Our method outperforms \textit{Adaptive BC} on almost all of the modified environments and is more stable with less variance in performance. This demonstrates that our method can adapt to different changes, even if the changes did not happen at training.  

\section{Conclusion}
In this paper, we propose the novel problem of adaptive offline-to-online RL where the dynamics can change at each episode during online fine-tuning. We present a residual learning framework with context encoding to train an adaptive policy. In contrast to previous offline-to-online RL approaches, our method can compensate for dynamics changes by considering a short history of state transitions from the environment. Moreover, our experiments demonstrated that it can generalize to out-of-distribution dynamics that were not present in the online fine-tuning stage. 

\textbf{Future work}
We believe that our method can be improved in multiple ways. For instance, we use a constant coefficient for balancing the importance of the offline policy and the adaptive residual policy. However, automatically tuning this hyper-parameter could alleviate the drop in performance at the early stages of training. 

\acks{
We acknowledge CSC – IT Center for Science, Finland, for awarding this project
access to the LUMI supercomputer, owned by the EuroHPC Joint Undertaking, hosted by CSC
(Finland) and the LUMI consortium through CSC. We acknowledge the computational resources
provided by the Aalto Science-IT project. 
J.~Pajarinen was partly supported by Research Council of Finland (345521). M.~Nakhaei was supported by Business Finland (BIOND4.0 - Data Driven Control for Bioprocesses).
A.~Scannell was supported by the Research Council of Finland from the Flagship program: Finnish Center for Artificial Intelligence (FCAI). 
}

\bibliography{reference.bib}

\end{document}